\documentclass{article}

\usepackage[final]{neurips_2024}

\usepackage[utf8]{inputenc} 
\usepackage[T1]{fontenc}    
\usepackage{hyperref}       
\usepackage{url}            
\usepackage{booktabs}       
\usepackage{amsfonts}       
\usepackage{nicefrac}       
\usepackage{microtype}      
\usepackage{xcolor}         
\usepackage{graphicx}
\usepackage{array}

\title{Can Vision-Language Models Replace Human Annotators: A Case Study with CelebA Dataset}

\author{%
  Haoming Lu \\
  Picsart AI Research \\
  Picsart Inc. \\
  \texttt{haoming.lu@picsart.com}  \\  
  \And
  Feifei Zhong \\
  College of Computing \\
  Georgia Institute of Technology \\
  \texttt{fzhong30@gatech.edu} \\
}

\newcommand{\shortcite}[1]{[\citenum{#1}]}

\setcitestyle{citesep={,}}

\begin{document}

\maketitle

\begin{abstract}
This study evaluates the capability of Vision-Language Models (VLMs) in image data annotation by comparing their performance on the CelebA dataset in terms of quality and cost-effectiveness against manual annotation. Annotations from the state-of-the-art LLaVA-NeXT model on 1000 CelebA images are in 79.5\% agreement with the original human annotations. Incorporating re-annotations of disagreed cases into a majority vote boosts AI annotation consistency to 89.1\% and even higher for more objective labels. Cost assessments demonstrate that AI annotation significantly reduces expenditures compared to traditional manual methods—representing less than 1\% of the costs for manual annotation in the CelebA dataset. These findings support the potential of VLMs as a viable, cost-effective alternative for specific annotation tasks, reducing both financial burden and ethical concerns associated with large-scale manual data annotation. The AI annotations and re-annotations utilized in this study are available on \href{https://github.com/evev2024/EVEV2024_CelebA}{GitHub}.

\end{abstract}

\vspace{-10pt}
\section{Introduction}

High-quality annotated data is recognized as a pivotal factor in the advancement of deep learning \shortcite{emam2021state, rasmussen2022challenge}. Nevertheless, manual data annotation presents significant challenges in terms of cost and ethical considerations \shortcite{denton2021whose}. Recent evolution in large language models (LLMs) \shortcite{zhao2023survey} has sparked enormous interest in their application to annotation and generation of text datasets \shortcite{tan2024large, bertaglia2023closing, acharya2023llm, yu2024large, xiang2022asdot}. Meanwhile, there are limited efforts \shortcite{zhao2022exploiting, han2024towards} exploring the capabilities of vision language models (VLMs) \shortcite{10445007} in processing unlabeled image data. Previous studies have verified that VLMs can create various types of annotations on raw image data. However, comprehensive evaluations of their annotation quality and cost-effectiveness are essential to assess their potential to replace manual annotation. \par

In this paper, we evaluated the capability of AI-driven image data annotation by comparing the quality and costs between manual annotations and annotations generated by one of the SOTA VLMs (LLaVA-NeXT \shortcite{liu2024llavanext}) on the CelebA \shortcite{liu2015faceattributes} dataset.
Overall, the contributions of this work include:
\begin{itemize}
    \item We confirmed that for specific image classification tasks, AI models can achieve performance comparable to human annotators, particularly excelling on more objective labels.
    \item Based on the similar quality and significantly lower costs, we argue that AI models already possess the potential to replace human annotation within certain scopes.
\end{itemize}

\section{Background}
\textbf{CelebFaces Attributes Dataset (CelebA) } \shortcite{liu2015faceattributes} is a public face attributes dataset with over 200K celebrity images, each with 40 binary attribute annotations. Although the dataset was manually created, analysis \shortcite{wu2023consistency} has pointed out errors and inconsistencies in the annotations. \par
\textbf{LLaVA-NeXT} \shortcite{liu2024llavanext} is an open-sourced SOTA multimodal model that enhances visual reasoning and OCR capabilities compared to LLaVA-1.5 \shortcite{liu2023llava}, which served as the foundation of many comprehensive studies of data, model, and capabilities of large multimodal models (LMM).

\section{Experiments}
\vspace{-10pt}
\begin{figure}[h]
  \centering
  \includegraphics[width=397pt]{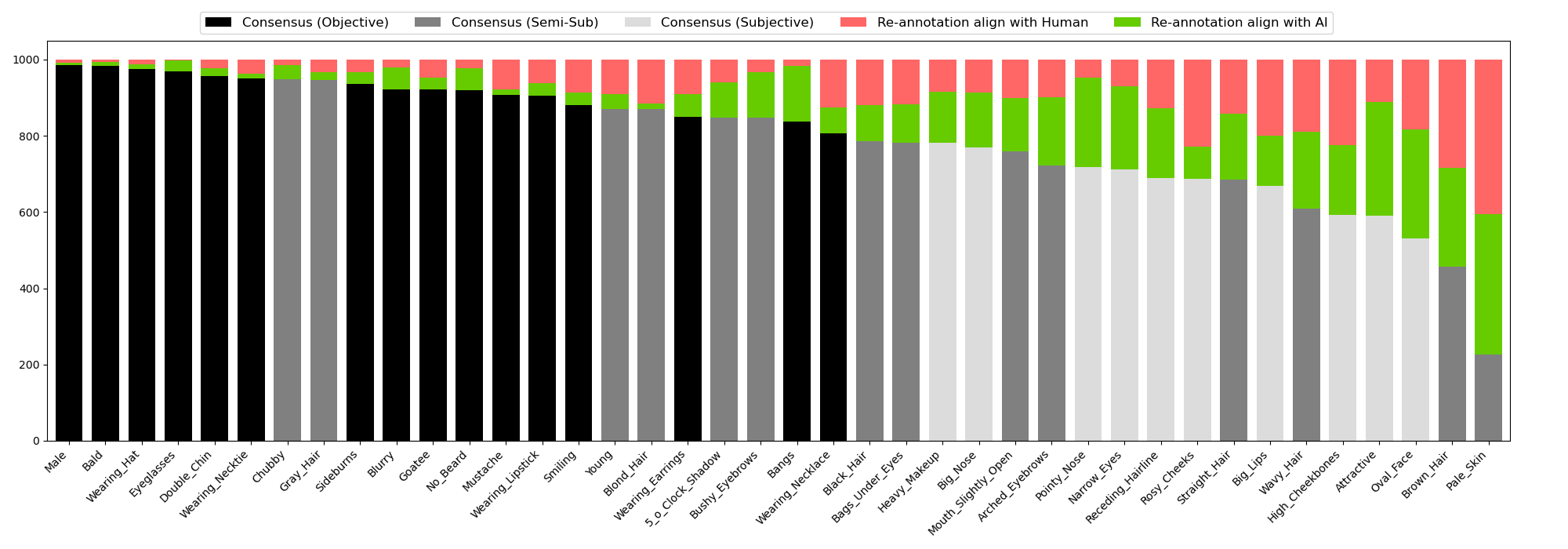}
  \vspace{-20pt}
  \caption{The performance of AI-generated annotations on the 40 CelebA attributes: (1) The more objective the attribute, the higher the agreement between AI and human annotations; (2) In cases of discrepancy, re-annotation aligned more with AI in exactly 50\% of the attributes.}
  \vspace{-5pt}
  \label{fig:overall}
\end{figure}
For each of the 40 binary attributes in the CelebA dataset, we designed a question that required the model to respond with only \textbf{yes} or \textbf{no}. For instance, the question posed for the attribute \textbf{Eyeglasses} was: \textit{Is this person wearing eyeglasses? Answer with only yes or no.} Annotations were generated on a randomly selected subset of 1000 images. Two reviewers re-annotated the attributes where AI and manual annotations differed. Inspired by \shortcite{wu2023consistency}, the attributes were divided into 3 groups based on their levels of objectivity. As illustrated in Figure \ref{fig:overall}, AI and the original human annotation reached \textbf{79.5\%} overall consensus, with higher numbers on more objective attributes. Conducting a majority vote with the additional re-annotation further boosts the match rate between AI annotations and the final consensus to \textbf{89.1\%}. Specifically, the re-annotation showed an equal preference for AI and human annotations, with each method favored more in exactly 20 attributes. \par

In our experiment, 40K labels for 1000 images were generated by a single NVIDIA RTX A6000 GPU at float16 precision using the HuggingFace implementation of the LLaVA-NeXT-8B model \shortcite{hfllavanext}. However, considering that in real life, the unit price of manual annotation decreases with scale, to fairly compare the costs of AI and human annotation, the cost estimation is based on constructing the entire CelebA dataset—generating 40 labels on each of the 200K images. The pricing for manual annotation is calculated based on publicly available quotes, whereas the cost of AI annotations is estimated from the average inference time per label and prices of renting GPU instances from Lambda \shortcite{lambda}. As shown in Table \ref{tab:comparison}, for image classification tasks like those in the CelebA dataset, AI annotation costs can be less than \textbf{1\%} of human annotation costs.

\begin{table}[h]
\centering
\begin{tabular}{lccccccc}
\toprule
& & \multicolumn{3}{c}{Human} & & \multicolumn{2}{c}{AI} \\ \midrule
Provider & & Google \shortcite{gcpquota} & AWS \shortcite{awsquota} & Scale AI \shortcite{scaleai} & & A6000 & A100 \\
\midrule
Cost (USD) & & 1100.00 & 530.00 & 2420.00 & & 4.36 & \textbf{4.01} \\ \bottomrule
\end{tabular}
\vspace{5pt}
\caption{The costs of manual and AI annotations (per 40K labels for 1000 images).}
\label{tab:comparison}
\end{table}
\vspace{-18pt}

\section{Conclusions and Limitations}
In this study, we evaluated the differences in quality and costs between the CelebA dataset constructed using AI models and the one manually created. Given the comparable quality and substantial cost advantages, Vision-Language Models (VLMs) have shown the potential to replace one or more annotation sources in scenarios where multiple annotations are utilized to enhance accuracy. Limited by the experiment's scale, the results may have been influenced by the biases of the AI model and the annotators' subjectivity. Future research could focus on whether VLMs can effectively handle more complex tasks and whether better interactions with AI models can improve the quality of annotations.

{
\small
    \bibliographystyle{ieeenat_fullname}
    \setcitestyle{numbers}
    \bibliography{reference}
}


\appendix

\section{Grouping of CelebA attributes}
As shown in Table \ref{tab:grouping}, the 40 attributes in CelebA dataset are divided into three groups:
\begin{itemize}
    \item Objective: These attributes are based on clear traits or easily identifiable elements, and are not open to personal interpretation. For example, whether the person is wearing glasses.
    \item Semi-Subjective: The attributes are based on ambiguous or less recognizable features that may be interpreted differently between observers. For instance, although youthfulness is a relatively objective attribute, people may have different perspectives on specific individuals.
    \item Subjective: These attributes are based on personal judgment and can vary significantly between observers. For example, whether a person is considered attractive entirely depends on the observer's preferences.
\end{itemize}

\begin{table}[h]
\centering
\label{tab:grouping}
\begin{tabular}{p{2.5cm}p{10.5cm}}
\toprule
\textbf{Group} & \textbf{Attributes} \\
\midrule
Objective & Male, Bald, Wearing\_Hat, Eyeglasses, Double\_Chin, Wearing\_Necktie, Sideburns, Blurry, Goatee, No\_Beard, Mustache, Wearing\_Lipstick, Smiling, Wearing\_Earrings, Bangs, Wearing\_Necklace \\
Semi-Subjective & Chubby, Gray\_Hair, Young, Blond\_Hair, 5\_o\_Clock\_Shadow, Bushy\_Eyebrows, Black\_Hair, Bags\_Under\_Eyes, Mouth\_Slightly\_Open, Arched\_Eyebrows, Straight\_Hair, Wavy\_Hair, Brown\_Hair, Pale\_Skin \\
Subjective & Heavy\_Makeup, Big\_Nose, Pointy\_Nose, Narrow\_Eyes, Receding\_Hairline, Rosy\_Cheeks, Big\_Lips, High\_Cheekbones, Attractive, Oval\_Face \\
\bottomrule
\end{tabular}
\vspace{3pt}
\caption{Grouping of the 40 CelebA attributes.}
\end{table}
Despite the minor differences in the grouping used between our study and that of \shortcite{wu2023consistency}, the choice of grouping does not affect the conclusions.

\section{Detailed comparison of AI annotation, original human annotation, and re-annotation}

Table \ref{tab:data} contains the detailed information that is illustrated in Figure \ref{fig:overall}. Column \textbf{AI=Human} is the number of images where the AI annotation is identical to the original human annotation; column \textbf{Re=AI} and \textbf{Re=Human} indicate the number of images where the re-annotation agrees with the AI annotation and the original human annotation respectively.

\begin{table}[h]
\centering
\begin{tabular}{lccc}
\toprule
Attribute & AI=Human & Re=AI & Re=Human \\
\midrule
Male & 986 & 8 & 6 \\
Bald & 983 & 7 & 10 \\
Wearing\_Hat & 976 & 12 & 12 \\
Eyeglasses & 970 & 2 & 28 \\
Double\_Chin & 957 & 23 & 20 \\
Wearing\_Necktie & 950 & 37 & 13 \\
Chubby & 949 & 14 & 37 \\
Gray\_Hair & 947 & 34 & 19 \\
Sideburns & 937 & 34 & 29 \\
Blurry & 922 & 21 & 57 \\
Goatee & 922 & 47 & 31 \\
No\_Beard & 919 & 22 & 59 \\
Mustache & 907 & 79 & 14 \\
Wearing\_Lipstick & 906 & 61 & 33 \\
Smiling & 881 & 87 & 32 \\
Young & 871 & 91 & 38 \\
Blond\_Hair & 870 & 116 & 14 \\
Wearing\_Earrings & 850 & 90 & 60 \\
5\_o\_Clock\_Shadow & 848 & 60 & 92 \\
Bushy\_Eyebrows & 847 & 32 & 121 \\
Bangs & 838 & 17 & 145 \\
Wearing\_Necklace & 807 & 125 & 68 \\
Black\_Hair & 785 & 119 & 96 \\
Bags\_Under\_Eyes & 782 & 118 & 100 \\
Heavy\_Makeup & 782 & 84 & 134 \\
Big\_Nose & 770 & 87 & 143 \\
Mouth\_Slightly\_Open & 759 & 100 & 141 \\
Arched\_Eyebrows & 722 & 98 & 180 \\
Pointy\_Nose & 717 & 48 & 235 \\
Narrow\_Eyes & 712 & 70 & 218 \\
Receding\_Hairline & 689 & 128 & 183 \\
Rosy\_Cheeks & 687 & 228 & 85 \\
Straight\_Hair & 686 & 143 & 171 \\
Big\_Lips & 669 & 199 & 132 \\
Wavy\_Hair & 608 & 190 & 202 \\
High\_Cheekbones & 592 & 224 & 184 \\
Attractive & 591 & 111 & 298 \\
Oval\_Face & 531 & 183 & 286 \\
Brown\_Hair & 456 & 284 & 260 \\
Pale\_Skin & 226 & 406 & 368 \\
\bottomrule
\end{tabular}
\vspace{5pt}
\caption{Detailed data underlying Figure \ref{fig:overall}.}
\label{tab:data}
\end{table}

\section{Calculation of AI annotation costs}
For inference with NVIDIA RTX A6000 GPU, the cost to annotate 1000 samples is estimated by 
$$0.49\times 40\times1000\times\frac{1}{3600}\times 0.8=4.36$$
where $0.49$ is the average inference time (seconds) per label, $40$ and $1000$ are the number of attributes per image and number of images, and $0.8$ is the price per hour according to Lambda \shortcite{lambda}. \par
Similarly, for A100, the cost is
$$0.28\times 40\times1000\times\frac{1}{3600}\times 1.29=4.01$$

\clearpage
\newpage

\section{Confidence intervals of the agreement percentages}

Below are the agreement percentages between AI annotation and human re-annotation / majority votes by different label categories (subject, semi-subjective, objective).

\begin{table}[h]
\centering
\begin{tabular}{lccc}
\toprule
Category & 90\% & 95\% & 99\% \\
\midrule
Objective & (0.8960, 0.9428) & (0.8910, 0.9479) & (0.8800, 0.9588) \\
Semi-Subjective & (0.6457, 0.8338) & (0.6250, 0.8545) & (0.5797, 0.8997)\\
Subjective & (0.6272, 0.7208) & (0.6163, 0.7317) & (0.5910, 0.7570) \\
\midrule
Overall & (0.7514, 0.8389) & (0.7427, 0.8477) & (0.7249, 0.8655) \\
\bottomrule
\vspace{3pt}
\end{tabular}
\label{conf1}
\caption{Confidence intervals of agreement between AI and human re-annotation by label categories.}
\end{table}

\begin{table}[h]
\centering
\begin{tabular}{lccc}
\toprule
Category & 90\% & 95\% & 99\% \\
\midrule
Objective & (0.9463, 0.9766) & (0.9430, 0.9799) & (0.9360, 0.9869) \\
Semi-Subjective & (0.8214, 0.9159) & (0.8110, 0.9263) & (0.7883, 0.9490) \\
Subjective & (0.7699, 0.8505) & (0.7605, 0.8599) & (0.7387, 0.8817) \\
\midrule
Overall & (0.8662, 0.9161) & (0.8611, 0.9211) & (0.8510, 0.9312) \\
\bottomrule
\vspace{3pt}
\end{tabular}
\label{conf2}
\caption{Confidence intervals of agreement between AI and majority votes by label categories.}
\end{table}

\end{document}